\documentclass[letterpaper]{article} 
\usepackage{aaai23}  
\usepackage{times}  
\usepackage{helvet}  
\usepackage{courier}  
\usepackage[hyphens]{url}  
\usepackage{graphicx} 
\usepackage{natbib}  
\usepackage{caption} 
\usepackage{algorithm}
\usepackage{algorithmic}

\usepackage{newfloat}
\usepackage{listings}
\DeclareCaptionStyle{ruled}{labelfont=normalfont,labelsep=colon,strut=off} 
\floatstyle{ruled}
\newfloat{listing}{tb}{lst}{}
\floatname{listing}{Listing}
\usepackage{microtype}
\usepackage{graphicx}
\usepackage{subfigure}
\usepackage{xcolor}

\usepackage{amsfonts}
 
\newcommand{\cA}{{\mathcal{A}}}

\newcommand{\cD}{{\mathcal{D}}}
\newcommand{\cS}{{\mathcal{S}}}

\newcommand{\cL}{{\mathcal{L}}}

\newcommand{\cT}{{\mathcal{T}}}

\newcommand{\cH}{{\mathcal{H}}}

\newcommand{\be}{{\textbf{e}}}

\newcommand{\bX}{\textbf{X}}

\newcommand{\bq}{\textbf{q}}
\newcommand{\bu}{\textbf{u}}

\newcommand{\bH}{\textbf{H}}

\newcommand{\bK}{\textbf{K}}

\newcommand{\br}{\textbf{r}}

\newcommand{\bpii}{\pmb{\pi}}

\newcommand{\bpsi}{{\pmb{\psi}}}

\newcommand{\bphi}{{\pmb{\phi}}}
\newcommand{\bmu}{{\pmb{\mu}}}
\newcommand{\btheta}{{\pmb{\theta}}}
\newcommand{\bdelta}{{\pmb{\delta}}}
\newcommand{\brho}{{\pmb{\rho}}}

\newcommand{\bbR}{\mathbb{R}}
\newcommand{\bbE}{\mathbb{E}}

\usepackage{mathtools}

\def\mtien#1{{\color{magenta}{#1}}}

\def\htien#1{}

\newcommand{\comments}[1]{{\color{blue}\textit{$\#$ #1}}}

\newcommand{\T}{\text{\tiny T}}

\newcommand{\KL}{\textsc{KL}}

\usepackage{amsmath}
\usepackage{amssymb}
\usepackage{mathtools}
\usepackage{amsthm}

\usepackage[capitalize,noabbrev]{cleveref}

\theoremstyle{plain}
\newtheorem{theorem}{Theorem}
\newtheorem{proposition}{Proposition}

\theoremstyle{definition}

\newtheorem{assumption}[theorem]{Assumption}
\theoremstyle{remark}

\title{Weighted Maximum Entropy Inverse Reinforcement Learning}
\author{
    The Viet Bui\textsuperscript{\rm 1}, Tien Mai \textsuperscript{\rm 1}, Patrick Jaillet \textsuperscript{\rm 2}
}
\affiliations{
    \textsuperscript{\rm 1}SCIS,
    Singapore Management University \\
        \textsuperscript{\rm 2}EECS,
    Massachusetts Institute of Technology\\

}

\usepackage{bibentry}

\begin{document}

\maketitle

\begin{abstract}
 We study  inverse reinforcement learning (IRL) and imitation learning (IM), the problems of recovering a reward or policy function from expert's demonstrated trajectories.  We propose a new way to improve the learning process by adding a weight function to the maximum entropy framework, with the motivation of having the ability to learn and recover  the stochasticity (or the bounded rationality) of the expert policy. Our framework and algorithms allow to learn both a reward (or policy) function and the structure of the entropy terms added to the Markov Decision Processes, thus enhancing the learning procedure. Our numerical experiments using  human and simulated demonstrations and with discrete and continuous IRL/IM tasks show that our approach  outperforms prior algorithms.
\end{abstract}

\section{Introduction}
Inverse reinforcement learning (IRL) and imitation learning (IM) \citep{russell1998learning,abbeel2004apprenticeship,Ziebart2008maximum} refer to  problems of learning and imitating expert's behavior by observing  their demonstrated trajectories of states and actions in some planning space. The experts are assumed to make actions by optimizing some accumulated rewards associated with states that they visit and the actions they make. The learner then aims at recovering such rewards or directly recovering expert's policies  to understand how decisions are made, and ultimately to imitate expert behavior.
The rationale behind IRL and IM is that
although a reward or a policy function might be a succinct and generalizable representation of an expert behavior, it is often difficult for the experts to provide such functions, as opposed to giving demonstrations.
So far, IRL and IM has been successfully applied in a wide range of problems such as  predicting driver route choice behavior \citep{Ziebart2008maximum} or planning for quadruped robots \citep{ratliff2009learning}.

Among available IRL and IM frameworks, maximum causal entropy IRL \citep{Ziebart2008maximum,ziebart2010_IRL_Causal} is a powerful probabilistic approach that has received a significant amount of attention over the decade.
 The framework can be built by adding entropy terms to Markov Decision Processes (MDP), which results in \textit{soft} and \textit{differentiable} policy structure. 
The main advantage of this framework is that it allows the removal of ambiguity between demonstrations and the expert policy and to cast the reward learning as a maximum likelihood estimation (MLE) problem. 
{However, a \textit{bottleneck} of the classical maximum causal entropy framework and IRL/IM algorithms based on it is that they seem to yield optimal policies having similar levels of stochasticity across states. In fact, in a real-life task, it is to be expected that the expert's policy at some states would be more (or less) random/stochastic than at other states.  For example, taxi drivers would be boundedly rational and his/her level of bounded rationality would not be the same over the whole network (e.g. they might be more or less rational in some areas, depending of the expected profit they can get or network density). 
On the  other hand, in a simulated task where the expert's policy is often assumed to be deterministic, the fact that the maximum entropy framework always yields random policies, leading to a mismatch between the modelled policy and the expert's one. Thus, a model/framework having the capability of adjusting the randomness of the policy would help diminish this inconsistency. We will demonstrate these points in more detail later in the paper.}

In this paper, we propose a way to enhance  maximum entropy IRL/IM algorithms by designing a new framework that allows to learn an additional function, beside a reward or policy function, that describes the stochasticity of the modeled policy. We do this by extending the maximum causal entropy framework \citep{ziebart2010_IRL_Causal}. Motivated by the fact that the  entropy terms added to the MDP have a fixed structure and cannot be learned, we generalize this structure by incorporating state feature information. Our aim is to maintain the softness property of the optimal policies from the maximum  entropy while allowing the structure of the  entropy function to be learned along with the reward/policy learning process.  This generalization leads to a weighted version of the maximum  entropy, leading to new IRL/IM algorithms being able to return both a reward/policy function and a weight vector for the entropy. Intuitively, an optimal policy  at each state is more deterministic if the weight value of this state is small, and more random when the weight is large. 
We provide \textit{equivalent constrained MDP formulations} and examples to demonstrate the benefits of being able to adjust the stochasticity of the policy over states. 
To the best of our knowledge, our algorithms are the first  to combine probabilistic reasoning about stochastic expert behavior with the ability of learning the stochasticity/randomness of the expert policies, allowing them  to outperform prior algorithms that only return a reward function or policy function.
Our weighted framework is \textbf{simple and general}, in the sense that it can be incorporated into any existing maximum entropy based IRL/IM algorithms.  
{We apply our weighted entropy framework into some popular IRL algorithms in the literature such as the  classical MaxEnt \citep{Ziebart2008maximum} and the Gaussian Process IRL \citep{Levine2011nonlinearIRL}.
We also develop weighted versions of the generative adversarial IRL and  IM algorithms \citep{Ho2016GAN_IRL,Fu2017Robust_IRL}, which are known to be  efficient to handle high-dimensional  continuous tasks. We provide experiments on discrete and continuous tasks, which demonstrate the advantages of our approach over prior IRL/IM algorithms.}  

\textbf{Related work:}
 Our algorithms  closely relate to prior IRL and IM algorithms proposed by  \cite{Levine2011nonlinearIRL};\cite{Ho2016GAN_IRL}; \cite{Jin2015inverse}; \cite{Finn2016Guided},  \cite{Fu2017Robust_IRL}. In particular,   the generative adversarial imitation learning (GAIL) algorithm proposed by \cite{Ho2016GAN_IRL} is a powerful approach that allows to learn a policy directly from demonstrations without recovering a reward function. Nevertheless, in many scenarios, a reward function (and a weight function in our approach) returned from IRL might be useful to infer expert's intentions or to avoid re-optimizing a reward/weight function in a new environment. \cite{Finn2016_connectionIRL} show a connection between generative adversarial networks (GAN) \citep{Goodfellow2014GAN}, maximum entropy IRL and energy-based models. They also propose the adversarial IRL (AIRL) framework that allows to learn a reward function based on the GAN idea. \cite{Fu2017Robust_IRL} develop an algorithm based on this AIRL framework, which provides a way to recover a reward function that is ``robust'' in different dynamic settings, and \cite{Yu2019multi} propose a multi-agent version of the AIRL.
 These generative adversarial algorithms are highly scalable in continuous domains  and are considered as state-of-the-art IRL/IM algorithms. 

\section{Background}
We start with a infinite-horizon Markov decision process (MDP) for an agent with finite states and actions, defined by the tuple $(\cS,\cA,\br,\bq,\gamma)$, where $\cS$ is a set of states $\cS = \{1,2,\ldots,|\cS|\}$, $\cA$ is a  set of actions, 
$\bq:\cS\times \cA\times\cS \rightarrow [0,1]$ is a transition probability function, i.e., $q(s'_t|a_t,s_t)$ is the probability of moving to state $s'_t\in\cS$ from $s_t\in \cS$ by performing action  $a_t\in \cA$ at time $t$, $\br$ is a vector of reward functions such that $r(a_t|s_t)$ is a reward function 
associated with state $s_t\in\cS$ and action $a_t\in\cA$ at time $t$, and $\gamma\in[0,1]$ is a discount factor. In the maximum causal entropy IRL framework \citep{ziebart2010_IRL_Causal}, we assume that the expert makes decisions by maximizing the entropy-regularized expected discounted reward (entropy-regularized MDP)
{\small \begin{align}
\max_{\bpii \in \Delta}\Bigg\{\bbE_{\tau \sim \bpii}\Bigg[\sum_{t=0}^{\infty}\gamma^t \Big( r(a_t|s_t) + \cH(\pi(a_t|s_t)) \Big)\Bigg]\Bigg\},\label{prob:Regularized-MDPs}
\end{align}}
where $\bpii$ is a policy function 
in which $\pi(a_t|s_t)$ is the probability of making action $a_t\in\cA$ at state $s_t\in\cA$ at time $t\in \{0,\ldots,\infty\}$
and $\Delta$ is the feasible set of $\bpii$ (i.e., $\Delta = \{\bpii|\; \sum_{a\in \cA} \pi(a|s) = 1, \;\forall a\in \cS\}$), 
 and $\cH(\pi(a_t|s_t)) = -\eta \ln \pi(a_t|s_t)$ are the entropy terms. 
The above MDP problem yields soft policies, which is essential to cast the IRL procedure as MLE. That is, one can define a parameterized structure for the reward function $r_{\btheta}(a|s)$ and infer the parameters $\btheta$ by solving the maximum likelihood problem 
$   \max_{\btheta} \left\{\bbE_{\tau \sim \cD^{E}}\left[ \ln P(\tau|\bpii^*) \right]\right\},$ 
where $\cD^{E}$ is the set of demonstrated trajectories, $\bpii^*$ is an optimal policy of \eqref{prob:Regularized-MDPs}, and   $P(\tau|\bpii^*)$ is the probability of trajectory $\tau$ under policy $\bpii^*$. In other words, we aim at seeking a reward function that maximizes the likelihood of expert's policy, assuming that the expert's policy is given by the entropy-regularized MDP. The stochasticity/randomness of the modelled policy is driven by parameter $\eta$. If $\eta \rightarrow \infty$, such a policy will tends to a random walk, and if $\eta \rightarrow 0$, then a policy will approach a deterministic one. Thus, we can adjust the stochasticity of the policy by modifying $\eta$. However, this will equally affect all the state policies. This is the reason why we say that the modelled policy seems to have  \textit{similar levels of stochasticity} across states.  

Proposition \ref{prop:p1} below shows that the entropy-regularized MDP (or maximum causal entropy) can be viewed  as a constrained MDP problem, where the objective is to maximize a standard (unregularized) expected reward while imposing an upper bound on the expected Kullback–Leibler (KL) divergence  between the desired policy and a random-walk policy ($\be/|\cA|$). Intuitively, by imposing a constraint on the expected KL divergence, we can make the desired policy closer to the random-walk one.
Clearly, if $\alpha \rightarrow 0$, then the desired policy should approach the random-walk one, and if $\alpha \rightarrow \infty$, the problem becomes unconstrained and the desired policy will approach a deterministic one. 

\begin{proposition}[Constrained MDP reformulation]
\label{prop:p1}
There is a scalar $\alpha \geq 0 $ such that the MDP problem \eqref{prob:Regularized-MDPs} can be formulated equivalently as
\begin{align}\small
\underset{\bpii}{\text{max}}\; &  \bbE_{\tau \sim \bpii}\Bigg[\sum_{t=0}^{\infty}\gamma^{t}  r(a_t|s_t) \Bigg] &\label{prob:general-uncertainty-model-s-rect} \\
\text{s.t.}\;  &  \bbE_{\tau \sim \bpii}\Bigg[\sum_{t=0}^{\infty}\gamma^{t} \eta  \KL\left(\pi(\cdot|s_t)\Big|\Big| \frac{\be}{|\cA|} \right) \Bigg] \leq \alpha,&\label{eq:constraint-10}
\end{align}
where $\be$ is a unit vector of size $|\cA|$ and $\KL(\cdot||\cdot)$ stands for the KL divergence between two discrete distributions. 
\end{proposition}
 In Constraint \eqref{eq:constraint-10}, 
  the term $\KL\left(\pi(\cdot|s_t)\Big|\Big| \frac{\be}{|\cA|} \right)$ measures the closeness between the policy at state $s_t$ and the random-walk policy.   
 In fact, one might expect that at some states, the policy would be more (or less) random/stochastic than policy at other states. However, the   expected KL divergence in \eqref{eq:constraint-10}  has a fixed structure, thus the stochasticity/randomness of the policy cannot be properly controlled and learned. In the following section, we present our  framework that allows to add weights to the expected KL divergence in \eqref{eq:constraint-10}, leading to a more flexible maximum entropy framework for IRL and IM.   

\section{Weighted  Entropy} 
Our aim here is to generalize the entropy function $-\eta \ln(\pi(a_t|s_t))$, in such a way that  (i) it can include  some state-action features, so its structure  can be learned together with the reward function, and (ii) the Markov problem still yields soft optimal policies, which is convenient for the reward learning procedure. To this end, let us consider a regularized  MDP with a weighted regularizer as $-\mu(s) \ln(\pi(a_t|s_t))$, where $\mu(s)$ is a weight function that can contain some feature information of state $s$ and can be  learned together with the rewards $r(a|s)$.
The Markov problem can be formulated  as 
{\small\begin{align}
\max_{\substack{\bpii}}\Bigg\{\bbE_{\tau \sim \bpii}\Bigg[\sum_{t=0}^{\infty}\gamma^{t} \Big( r(a_t|s_t)  - \mu(s_t)\ln(\pi(a_t|s_t)) \Big)\Bigg]\Bigg\}.\label{prob:General-Regularized-MDPs-2}
\end{align}}
Since the generalized formulation contains a weight function $\mu(s_t)$ associated with each entropy term $\ln \pi(a_t|s_t)$, we call this MDP framework as \textbf{Weighted Entropy}. Note that the classical maximum entropy is just a\textit{ special case} of the formulation above where $\mu(s)$ are the same over states. 
In the following, we investigate some properties of the generalized framework. In analogy to Proposition \ref{prop:p1}, in Proposition \ref{prop:prop-3} below we also formulate \eqref{prob:General-Regularized-MDPs-2} as a constrained MDP problem.
\begin{proposition}[\textit{Constrained MDP reformulation}]
\label{prop:prop-3}
There is a scalar $\alpha\geq 0$ such that \eqref{prob:General-Regularized-MDPs-2} can be formulated equivalently as 
\small{\begin{align}
\underset{\bpii}{\text{max}}\; &  \bbE_{\tau \sim \bpii}\Bigg[\sum_{t=0}^{\infty}\gamma^{t}  r'(a_t|s_t) \Bigg] &\nonumber \\
\text{s.t.} \; &  \bbE_{\tau \sim \bpii}\Bigg[\sum_{t=0}^{\infty}\gamma^{t}  \mu(s_t)\KL\left(\pi(\cdot|s_t)\Big|\Big| \frac{\be}{|\cA|} \right) \Bigg] \leq \alpha,&\label{eq:constraint-11}
\end{align}}
where $r'(a_t|s_t) = r(a_t|s_t) + \ln |\cA| \mu(s_t)$. 
\end{proposition}
In Constraint \ref{eq:constraint-11}, each KL divergence is multiplied by a state-wise weight $\mu(s)$. Intuitively, if $\mu(s)$ takes a high value, then the corresponding KL divergence  at state $s$ should be small  and the policy $\pi(\cdot|s)$ should be close to the random-walk (i.e., more random), and if $\mu(s)$ takes a small value, the KL divergence value should be larger and the policy $\pi(\cdot|s)$ should be closer to a deterministic one. In other words, the weights $\mu(s)$ can affect the randomness of the optimal policy. 
In IRL, one can define $\mu(s)$ as a function of the feature information, allowing these weights to be learned together with the reward function $r(a|s)$.

To illustrate the benefits from having a  model being flexible in learning the stochasticity of the policy over states, we give the following real-life example. In a transportation network, one would be interested in  the problem of learning from  taxi-drivers' behavior, where states are defined as intersections in the network, and the aim is to travel around the network to find customers. Typically, states with  high rewards often stick together (e.g., in downtown areas). This  defines low- and high-reward areas in the map. Intuitively, it is expected that drivers would be bounded rational and the levels of bounded rationality would be different across the network,
depending on, for instance, the distribution of the rewards over the network or network densities. This hidden information can be partially  recovered by the our weighted maximum entropy framework by learning, in addition to the reward/policy function, the weight function $\mu(s)$. 


In analogy  to the maximum causal entropy framework \citep{ziebart2010_IRL_Causal},  it is straightforward to see that a solution to the Markov problem \eqref{prob:General-Regularized-MDPs-2}  solves the following maximum \textbf{weighted casual entropy} problem
\begin{align}
\underset{\bpii}{\text{max}}\qquad & - \bbE_{\tau \sim \bpii}\Big[\sum_{t=0}^{\infty}\gamma^{t} \mu(s_t) \ln(\pi(a_t|s_1))  \Big] & \nonumber \\
\text{s.t.} \qquad &  \bbE_{\tau \sim \bpii}\Big[R(\tau) \Big]  = \bbE_{\tau\sim \bpii_E} \Big[R(\tau)\Big],& \nonumber
\end{align}
where $R(\tau) = \sum_{t=0}^{\infty}\gamma^{t} r(a_t|s_t)$, $\bpii_E$ is the expert's policy and $\bbE_{\tau\sim \bpii_E} \Big[R(\tau)\Big]$ is the expected accumulated reward under expert's policy.
Note that in our formulation,  the causal entropy function $\bH(\bpii,\bmu) = -\bbE_{\tau \sim \bpii}\Big[\sum_{t=0}^{\infty}\gamma^{t} \mu(s_t) \ln(\pi(a_t|s_1))]$ no-longer has a fixed structure as in \cite{ziebart2010_IRL_Causal}. Instead, the structure of this function depends on $\mu(s)$ and  can be learned during the IRL procedure.  

It can be also seen that a solution to the above maximum weighted causal entropy problem also minimizes the  \textbf{worst-case prediction weighted log-loss}
\[
\min_{{\brho \in\Delta}} \max_{\substack{\bpii\in\Delta \\ \bbE_{\bpii}[R(\tau)]= \bbE_{\bpii_E}[R(\tau)]}} \left\{-\sum_{t=0}^\infty \gamma^{t} \mu(s_t)\ln \rho(a_t|s_t) \right\}. 
\]	
In other words, the  maximum weighted causal entropy can be viewed as a zero-sum game where the adversary chooses a distribution over actions/states to maximize the predictor’s \textit{weighted} log-loss value, and the predictor tries to choose a distribution to minimize it.

We now turn our attention to the question of how to compute an optimal policy of  \eqref{prob:General-Regularized-MDPs-2}, which is essential for our IRL algorithms. As a standard way to solve an infinite-horizon Markov problem, we define a value function as the expected regularized reward from any state $s\in\cS$
\begin{align}
 V^*(s) = &\max_{\substack{\bpii}}\Bigg\{\bbE_{\tau \sim \bpii}\Bigg[\sum_{t=0}^{\infty}\gamma^{t} \Big( r(a_t|s_t) \nonumber\\
 &- \mu(s_t)\ln(\pi(a_t|s_t)) \Big)\Bigg| s_0=s\Bigg]\Bigg\}. \nonumber
\end{align}
Then, similar to prior work \citep{Bloem2014infinite}, we can show that $V^*$ is a solution to a contraction mapping, which guarantees the existence and uniqueness of $V^*$ for any reward and weight functions. Moreover, a soft policy is optimal to the Markov problem. We summarize our main claim in Theorem \ref{th:optimal-policy} below. 
\begin{theorem}[Optimal policy]
\label{th:optimal-policy}
The policy defined below is optimal to the Markov problem \eqref{prob:General-Regularized-MDPs-2}
\begin{equation*}
\pi^*(a|s) = \frac{\exp(Q(a,s,V^*)/\mu(s))}{\sum_{a'} \exp(Q(a',s,V^*)/\mu(s))},
\end{equation*}
where $Q(a,s,V^*) = r(a|s)+\gamma \bbE_{s'}[V^*(s')]$ and $V^*$ is a solution to the contraction mapping $\cT[V] = V$, where
$
    \cT[V] = \mu(s) \ln \left(\sum_{a} \exp(Q(a,s,V)/\mu(s))\right).
$
\end{theorem}
The proof is given in the supplement. One of the key advantages of the above soft formulations is that if we define the reward $r(a|s)$ and the weights $\mu(s)$ as differentiable functions (e.g., neutral networks) of the feature information and some structure parameters, the value function and optimal policy are also differentiable, which allows to infer the structure parameters via a gradient-based optimization algorithm.  
\section{IRL and IM Algorithms}
We design new IRL and IM algorithms by incorporating our weighted entropy function into MLE-based and generative adversarial IRL/IM algorithms, recalling that MLE-based algorithms \citep{Ziebart2008maximum,Levine2011nonlinearIRL} are more  classical and often work well with low-dimensional tasks while the generative adversarial ones \citep{Fu2017Robust_IRL,Ho2016GAN_IRL} would  be more efficient to handle large-scale continuous tasks. 

\subsection{MLE-based IRL}
In this context the structural parameters are inferred by maximizing the log-likelihood of the expert's demonstrations.
The maximum likelihood problem can be written as
$
\max_{\btheta,\bpsi}\left\{\cL(\cD^E|\btheta,\bpsi) = \sum_{\tau \in\cD^E}  \ln P(\tau| \bpii^*_{\btheta,\bpsi} )  \right\},
$
where $\bpii^*_{\btheta,\bpsi}$ is the optimal policy of the entropy-regularized MDP problem \eqref{prob:General-Regularized-MDPs-2} with rewards $r_{\btheta}(a|s)$ and entropy regularizer $\cH^G(a|s) = -\mu_{\bpsi}(s)\ln \pi(a|s)$. 
If we denote by $V^{\btheta,\bpsi}$ the value function under rewards $r_{\btheta}(a|s)$ and weights $\mu_{\bpsi}(s)$, and $Q^{\btheta,\bpsi}(s,a) :|\cS|\times |\cA| \rightarrow \bbR$ as the Q-value function of the Markov problem \eqref{prob:General-Regularized-MDPs-2}, i.e.,  $Q^{\btheta,\bpsi}(s,a) = r_{\btheta}(a|s) +\gamma \bbE_{s'}[V^{\btheta,\bpsi} (s')]$, then the maximum likelihood problem  becomes
\begin{equation}
\label{prob:MLE-2}
\max_{\btheta,\bpsi}\left\{ \sum_{\tau \in\cD^E} \sum_{(s,a) \in\tau} \frac{\left(Q^{\btheta,\bpsi}(s,a) - V^{\btheta,\bpsi}(s) \right)}{\mu_{\bpsi}(s)}  \right\}.
\end{equation}
The gradients of $Q^{\btheta,\bpsi}(s,a)$, $V^{\btheta,\bpsi}(s)$ with respect to $\btheta$ and $\bpsi$ are necessary for an efficient IRL algorithm and we provide them in the supplement. 

The reward structure can be linear with respect to the feature information.
However, if the selected features do not form a good linear basis for the rewards, a nonlinear structure can achieve much better performance. 
Gaussian Process (GP) IRL is a state-of-the-art nonlinear structure in the context \citep{Levine2011nonlinearIRL, Jin2015inverse}. The idea is to use a GP to model and learn the reward function. 
That is, the rewards can be modelled as 
$\br = \bK^\T_{\br,\bu} \bK^{-1}_{\bu,\bu} \bu,$
where $\bu$ are structure parameters representing  the rewards associated with some  feature coordinates $\bX_\bu$, $\bK_{\bu,\bu}$ is a covariance matrix of the inducing point values $\bu$ located at $\bX_\bu$ and $\bK_{\br,\bu}$  is a covariance matrix of the rewards $\br$ with the inducing point
values $\bu$ \citep{Rasmussen2003gaussian}.  The entries of these covariance matrices are determined by a kernel function of hyper-parameters $\btheta$. So, we can write  $\br = \bK^\T_{\br,\bu}(\btheta) \bK^{-1}_{\bu,\bu}(\btheta) \bu$. Following \cite{Levine2011nonlinearIRL}, the log-likelihood function becomes
\begin{align}
    \cL(\cD^{E}|\bu,\btheta,\bpsi,\bX_\bu) &= \underbrace{\cL\left(\cD^{E}|\bpsi,\br = \bK^\T_{\br,\bu}(\btheta) \bK^{-1}_{\bu,\bu}(\btheta) \bu\right)}_{\text{IRL log-likelihood}}\nonumber\\
    &+ \underbrace{\ln P\left(\bu,\btheta|\bX_\bu\right)}_{\text{GP log-likelihood}},\nonumber
\end{align}
where $P\left(\bu,\btheta|\bX_\bu\right)$ is the GP marginal likelihood, which can be expressed as a function of the parameters $\bu$, covariance matrix $\bK_{\bu,\bu}$ and a hyper-parameters prior $P(\btheta)$. 
The gradients of the  \textit{IRL log-likelihood} with respect to $\bpsi, \btheta$ and $\bu$ can be obtained using the deviations provided in the supplement and the derivations of $\br$ with respect to $\bu,\btheta$, which can be found in  \cite{Levine2011nonlinearIRL}.

Algorithm \ref{algo:IRL} describe the  main phases of our IRL approach. In the first phase we try to shape a reward function and and in the second phase we train both the rewards and weights simultaneously. The algorithm requires  the gradients  of the likelihood function $\cL(\cD^E|\btheta,\bpsi)$ w.r.t. $\btheta$ and $\bpsi$, thus requires the gradients of the value function $V^*$, for which we provide an algorithm to compute the value function and its gradients in the supplements. The partial derivatives $ {\partial \cT[V]}/{\partial \psi_i}$ and $ {\partial \cT[V]}/{\partial \theta_i}$ as well as the formulations for the gradients of $\cL(\cD^E|\btheta,\bpsi)$ w.r.t. $\btheta$ and $\bpsi$ can be also found in the supplements.   

\begin{algorithm}[htb]
\caption{\textit{MLE-based IRL algorithm}}
\label{algo:IRL}
\begin{algorithmic}[1]
\STATE \comments{Phase 1: Shaping the rewards} 
\STATE Compute: 
$\overline{\btheta} = \text{argmax}_{\btheta} \left\{\cL(\cD^E|\btheta,\bpsi)\Big| \bpsi = \textbf{0}\right\}.$ 
\STATE \comments{Phase 2: Training the rewards and weights simultaneously}
\STATE Solve $\max_{\btheta,\bpsi} \{\cL(\cD^E|\btheta,\bpsi)\}$ using gradient decent with starting point $(\btheta^0,\bpsi^0) = (\overline{\btheta},\textbf{0})$.  
\end{algorithmic}
\end{algorithm}

\subsection{Generative Adversarial IRL/IM}
\label{sec:gan-based}
In AIRL \citep{Fu2017Robust_IRL}, one can train a discriminator $D(s,a,s')$ to classify expert data from policy samples  and update  the policy to confuse the a discriminator. At each step the algorithm, samples (or trajectories) are generate by executing the current policy $\bpii$ and the discriminator is trained via logistic regression to classify the expert data  and the sampled trajectories. The policy is then updated using the reward function $\widetilde{r}(s,a,s') = \ln D(s,a,s') - \ln (1-  D(s,a,s'))$. \citep{Fu2017Robust_IRL}  propose to use the discriminator of the form $D(s,a,s') = \exp(f(s,a,s'))/\left(\exp(f(s,a,s')) + \pi(a|s)\right)$, for all $s,s'\in\cS, a\in\cA$, where $f(s,a,s')$ is a disentangled reward function. In our context, to incorporate the weights $\mu(s)$ to the AIRL algorithm, we propose to use the following discriminator 
\[
D_{\btheta,\bdelta, \bpsi}(s,a) = \frac{\exp(f_{\btheta,\bdelta}(s,a,s'))}{ \exp(f_{\btheta,\bdelta}(s,a,s')) + \exp(\mu_{\bpsi}(s))  \pi(a|s))},
\]
where the policy $\pi(a|s)$ is multiplied by the weight $\exp(\mu_{\bpsi}(s))$. The disentangled reward is defined as $f_{\btheta,\bdelta}(s,a,s') = r_{\btheta}(a|s) +\gamma h_{\bdelta} (s') - h_{\bdelta}(s)$. Thus, at each step, the policy is updated by solving the Markov decision problem with rewards 
\[
\begin{aligned}
\widetilde{r}_{\btheta,\bdelta, \bpsi}(s,a,s') &= \ln D_{\btheta,\bdelta, \bpsi}(s,a,s')\\
&\qquad- \ln (1-  D_{\btheta,\bdelta, \bpsi}(s,a,s'))\\
&=f_{\btheta,\bdelta}(s,a,s') - \mu_{\bpsi}(s) \ln \pi(a|s).
\end{aligned}
\]
The use of the above discriminator then can be justified by seeing that the objective of  Markov problem with rewards $\widetilde{r}_{\btheta,\bdelta, \bpsi}(s,a,s')$ becomes
\[
\bbE_{\tau \sim \bpii}\Bigg[\sum_{t=0}^{\infty}\gamma^{t} \Big( f_{\btheta,\bdelta}(s_t,a_t,s_{t+1})  - \mu_{\bpsi}(s_t)\ln(\pi(a_t|s_t)) \Big)\Bigg],
\]
which is consistent with the weighted entropy formulation in \eqref{prob:General-Regularized-MDPs-2}.

The weighted AIRL is described in Algorithm \ref{algo:WAIRL} below.
\begin{algorithm}[htb]
\caption{\textit{Weighted AIRL}}
\label{algo:WAIRL}
\begin{algorithmic}[1]
\STATE Given expert trajectories $\tau^E_i$, $i=1,\ldots$\\
\STATE Initial policy $\bpii$  and discriminator $D_{\btheta,\bphi}(s,a) $
 \FOR{$t = 1,\ldots$} 
    \STATE Generate trajectories $\tau_i$ by executing the current policy $\bpii$\\
    \STATE Train $D_{\btheta,\bphi}(s,a)$ by binary logistic regression to classify expert data $\tau_i^E$ and the generated trajectories $\tau_i$, $i =1,\ldots$\\
    \STATE Compute rewards
    \[
     r'_{\btheta,\bphi}(a|s) = \ln D_{\btheta,\bphi}(s,a) - \ln (1-D_{\btheta,\bphi}(s,a))
    \]
    Update policy $\bpii$ using rewards $r'_{\btheta,\bphi}(a|s)$  and an policy optimization method.
 \ENDFOR
\end{algorithmic}
\end{algorithm}

The GAIL algorithm \citep{Ho2016GAN_IRL} runs in the same manner as AIRL, except that  the algorithm does not recover disentangled rewards but directly recover a policy function. To incorporate the weighted entropy to  GAIL, we simply use the weighted entropy function $\cH(a|s) = -\mu_{\bpsi}(s) \ln \pi_{\btheta}(a|s)$, noting that $\btheta$ now are the parameters of the policy $\bpii$.  

\section{Experiments}
\label{sec:exper}
We provide experiments with discrete and continuous control tasks. For the discrete ones, we compare MLE-based IRL algorithms, i.e., MaxEnt, Gaussian Process IRL \citep{Ziebart2008maximum,Levine2011nonlinearIRL}, with our weighted versions. For continuous tasks, for which generative adversarial  IRL/IM algorithms are known to be more useful,
we will compare AIRL and GAIL \citep{Fu2017Robust_IRL,Ho2016GAN_IRL} with  our weighted  versions. The experiments were conducted using  a PC with CPU Ryzen 9 3900X 12-core processor and 32GB RAM (no GPUs).

\subsection{Discrete control tasks}

We will use simulated (Objectworld and Highway Driving Behavior) and real-life environments (Driver Route Choice Behavior). We incorporate our weighted  framework into the classical MaxEnt \citep{Ziebart2008maximum}  and the GPIRL \citep{Levine2011nonlinearIRL}, which results in two new IRL algorithms denoted by W-MaxEnt and W-GPIRL, respectively. 
There are other IRL methods such as the FIRL, MMP, MWAL, MMPBoost and LEARCH \citep{Levine2010feature,Bagnell2007boosting,abbeel2004apprenticeship,Ratliff2006maximum,ratliff2009learning,syed2008game}, but they are outperformed by the MaxEnt and GPIRL \citep{Levine2011nonlinearIRL}. Thus, we only show comparison results for these two algorithms  and our weighted versions. 

For the real-life environment, since  the true rewards are unknown, we only evaluate how the IRL algorithms recover human's trajectories. For the simulated environments, similar to previous work \citep{Levine2011nonlinearIRL} we use the \textit{``expected value difference''} score, which measures how a learned policy performs under the true rewards. 
We evaluate the  IRL outputs  on both environments on which they were learned and random environments (denoted by ``\textit{transfer}''). For the latter, we bring the learned parameters of the reward and weight functions to compute rewards and optimal policies in the new environments. 
Each algorithm is evaluated with both continuous and discrete features and each test is repeated eight times with different  random environments. All the algorithms are implemented using the IRL toolbox provided by \cite{Levine2011nonlinearIRL}. We keep the same hyperparameters used in \cite{Levine2011nonlinearIRL}.   

\begin{figure}[htb] 
\centering
    \includegraphics[width=0.9\linewidth]{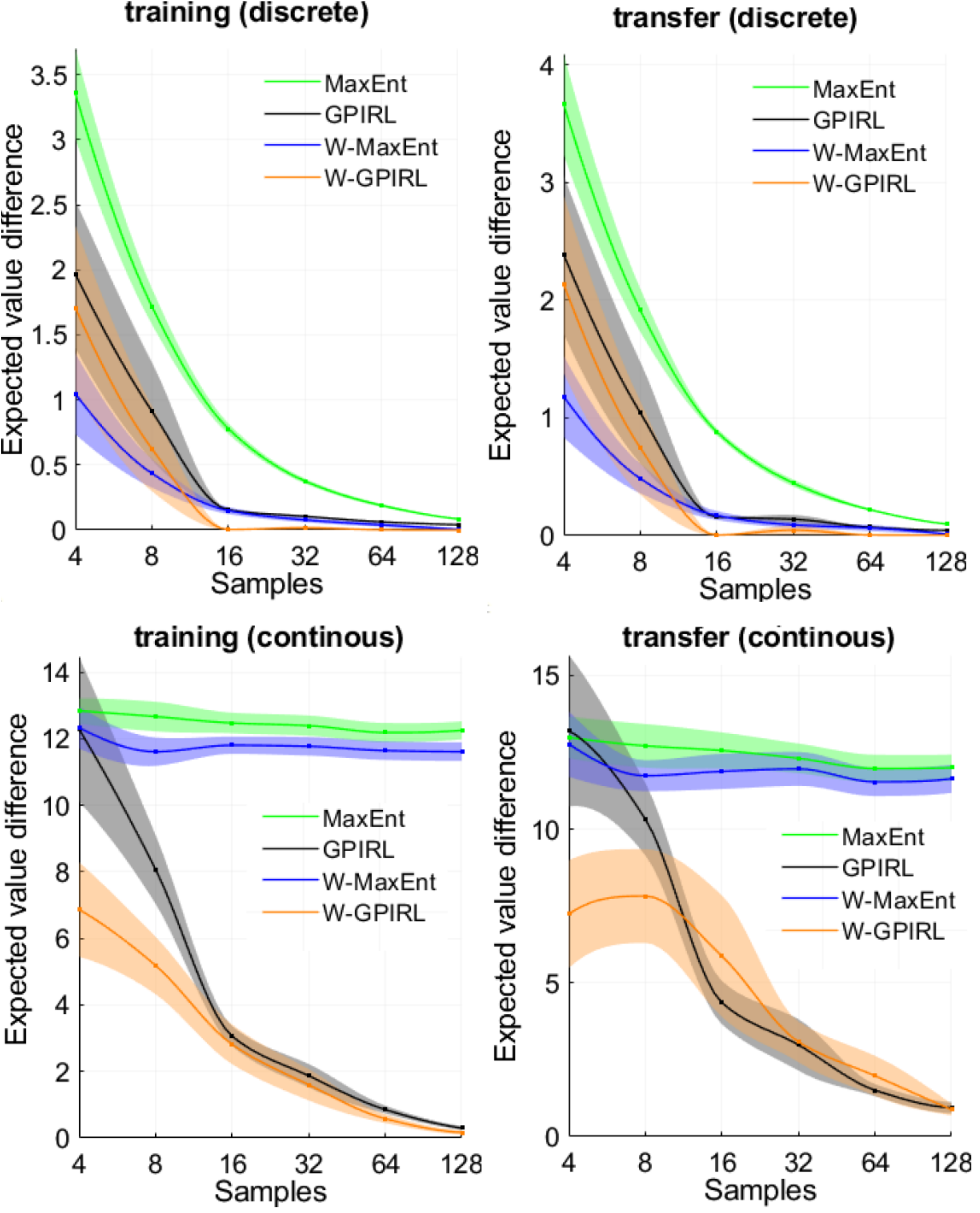} 
    \caption{\small Experiments with $32\times32$ objectworld, solid curves show the mean and the shading shows the standard errors.  } \label{fig:obj_expected_value} 
\end{figure}

\textbf{{Objectworld. }}
The Objectworld is an $N\times N$ grid of states in which objects are randomly placed. 
Each state has $2C$ basic continuous features. 
 For the discrete feature case, we discretize the continuous features, resulting in $2CN$ feature values.
Demonstrations are paths of length 8 generated by the true rewards.
 In fact, we keep the same  settings  as in \cite{Levine2011nonlinearIRL}.
 We run the experiments  with $N=32$, $C=2$ and the number of samples varies from 4 to 128. The means and standard errors of the scores are plotted in Fig.  \ref{fig:obj_expected_value}. Our algorithms (W-MaxEnt  and W-GPIRL) consistently and significantly outperform the other methods with 16 or less examples
 With more than 16 examples, the differences are smaller but our methods are still slightly better. 
 For the case of  discrete features and less than 32 examples, the W-MaxEnt outperforms other nonlinear-reward-structure methods (GPIRL, W-GPIRL). For the case of  continuous features, the GP-based methods perform much better than the MaxEnt-based ones with 8 or more samples. When the sample size is equal to or less than 8, while the performance of GPIRL and MaxEnt are comparable, they are substantially outperformed by our W-GPIRL algorithm. 

The experiments for the  Highway Driving Behavior task are provided in the supplement which also demonstrates that our weighted algorithms outperform their classical counterparts. 

\textbf{{Driver Route Choice Behavior with Human Demonstrations.}} 
The human dataset contains a total of 1832 trajectories of taxi drivers collected in  the network of Borlänge, Sweden. 
At each state, an action is to move to one of the connected next states with no uncertainty. 
This data was used in some  established route choice modeling studies \citep{Fosgerau2013_RL,Mai2015_nestedRL}. In this experiment, we only test the MaxEnt and W-MaxEnt algorithms because (i) the linear structure  is more useful interpreting how some important features (e.g., travel time) affect driver's behavior, (ii) the GP approach is two expensive, as the demonstrations contain many drivers' destinations (absorbing states) and for each destination we need to solve one MDP problem to get an optimal policy. Moreover, since we are not aware of the true rewards, we only evaluate how the algorithms recover drivers' trajectories. We do that by running the algorithm on a training set  and use the learned rewards/weights to generate trajectories. These trajectories are then compared to those in a test set to compute the ``average of maching'' and ``90\% matching'' scores  as in \cite{Ziebart2008maximum}.  
We place 80\% of the taxi trajectories into the training set and the remainder into the test set. We report the comparison results in Table \ref{tab:table_of_results_for_transport}. The first and second rows clearly show that the W-MaxEnt returns significantly larger log-likelihood values for both training and test sets. For both metrics on the third and fourth rows (Avg. Matching and 90\% Matching),  W-MaxEnt outperforms the MaxEnt. 

\begin{table}[htb]\small
\centering
\begin{tabular}{l|l|l}
  & MaxEnt     & W-MaxEnt\\ 
\hline
Log Prob. (training)& -2074.3 & \textbf{-1988.8}   \\ 
\hline
Log Prob. (test)    & -566.4  & \textbf{-523.4 }   \\ 
\hline\hline
Avg. Matching & 87.6\%  & \textbf{89.3}\%    \\ 
\hline
90\% Matching & 63.5\%  & \textbf{67.1\%}    \\ 
\end{tabular}
\caption{\small Comparison of the MaxEnt and  W-MaxEnt with human demonstrations.}
\label{tab:table_of_results_for_transport}
\end{table}
\subsection{Continuous Control Tasks}
We incorporate the weighted entropy into the AIRL and GAIL algorithms  and denote the new algorithms as WAIRL and WGAIL.
 Guide Cost Learning (GCL)  \citep{Finn2016_connectionIRL} is also a generative adversarial IRL algorithm  but we do not consider it here, as it is generally outperformed by the AIRL \citep{Fu2017Robust_IRL,Arnob2020off}.
We first compare the performances of these algorithms (AIRL, GAIL, WAIRL and WGAIL) with transfer learning tasks (the training  and testing environments are different) and then with some other high-dimensional imitation tasks. We use  the code published at \url{https://github.com/ku2482/gail-airl-ppo.pytorch} to implement and test the algorithms. All the hyperparameters are kept the same in our experiments.

\begin{figure}[htb] 
\centering
    \includegraphics[width=0.7\linewidth]{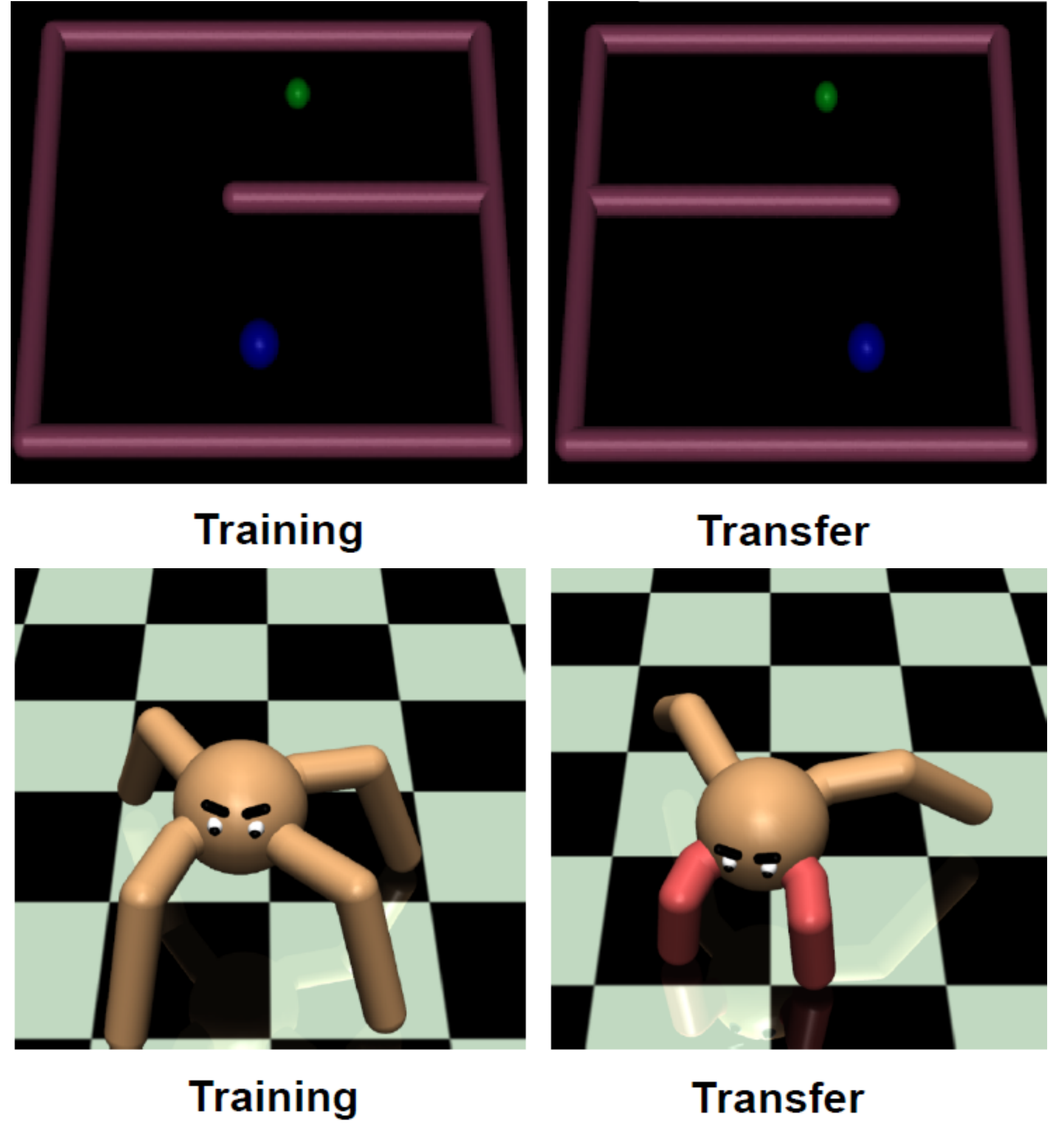} 
    \caption{\small Point Mass-Maze and Disabled-Ant for transfer learning.} 
    \label{fig:transfer-evn} 
\end{figure}

\textbf{Transfer learning.} 
We perform the experiment with two continuous control tasks that were designed previously for transfer learning. The first task (Point Mass-Maze) is to navigate to a goal position in a small maze when the position of wall is changed between the train and test times. The  second task (Ant-Disabled) involves  training a quadrupedal ant to run forwards  and in the test time two front legs of the ant are disabled. The tasks are shown in Fig. \ref{fig:transfer-evn} and more details can be found in  \citep{Fu2017Robust_IRL}.  We train  AIRL and WAIRL with state-only and state-action rewards, noting that the state-only model was shown to be better  in the context of AIRL \citep{Fu2017Robust_IRL}. 
To evaluate the performances the considered algorithms under transfer learning, reward/weight/policy functions are learned on the training environment, and then we bring these functions to the test environment to re-optimize the policy and compute scores (i.e., expected true rewards under re-optimized policies).
Table \ref{tab:transfer-tasks} reports the comparison results for the transfer learning experiments. The mean scores are reported over 5 runs (the higher the better). Our weighted algorithms (WAIRL and WGAIL) consistently outperform their counterparts (AIRL and GAIL).  WAIRL performs the best for Ant-Disabled. Previous work show that GAIL does not perform well for transfer learning tasks, but with the inclusion of the weight function we observe that it is significantly improved  and even outperforms AIRL and WAIRL on the Point Mass-Maze task.  

 \begin{figure}
\centering
  \centering
  \includegraphics[width=1.0\linewidth]{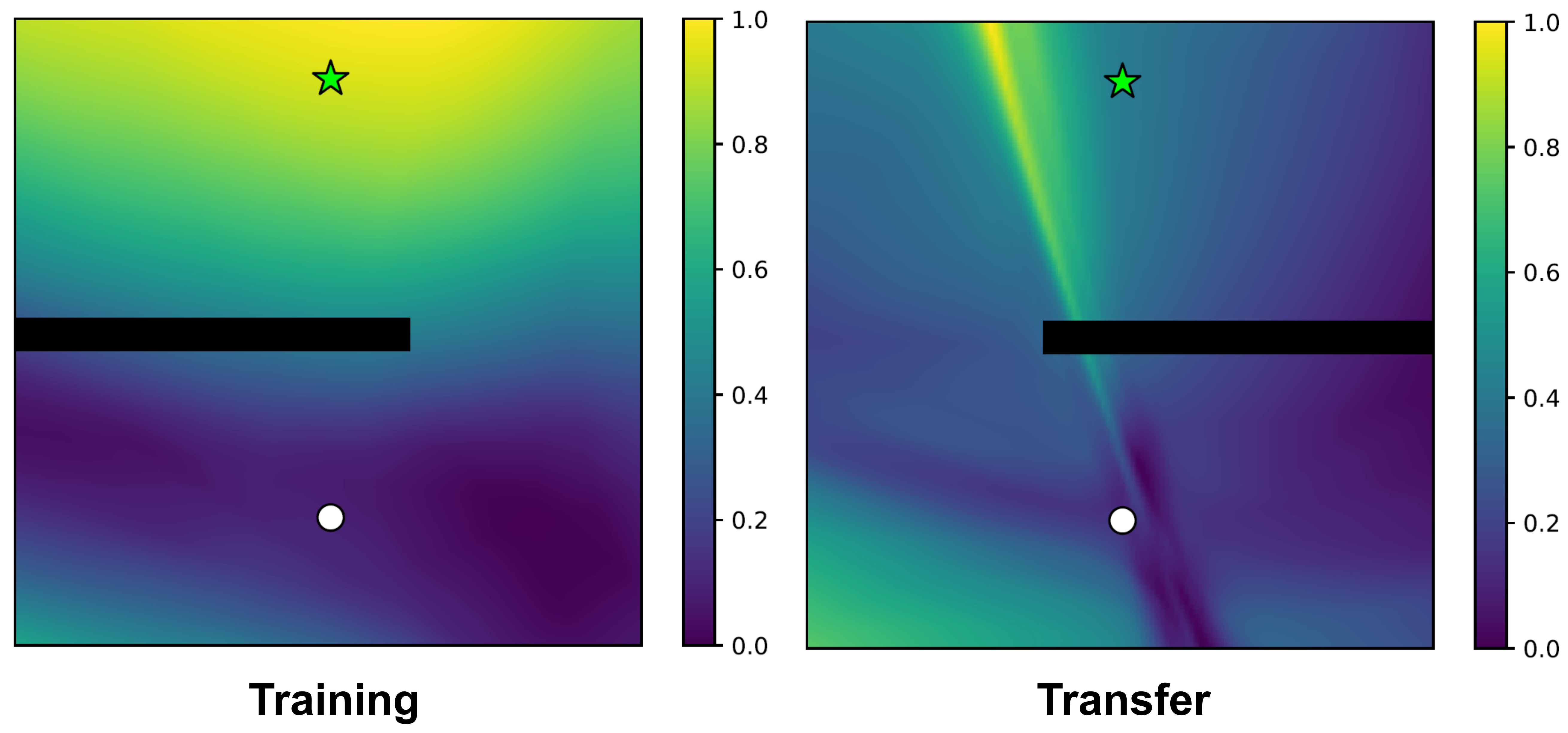}
\caption{Heatmaps of WAIRL-StateOnly's weights $\mu(s)$ for Point Mass-Maze task.}
\label{fig:mu}
\end{figure}

\begin{table}[htb]\small
\centering
\begin{tabular}{l|l|l}
Algorithms 	& Ant-Disabled 	& Point Mass-Maze\\
\hline
GAIL 	& 1.31$\pm$7.67 	& -24.37$\pm$11.82\\
\hline
WGAIL 	& 14.29$\pm$36.33 	& \textbf{-7.40$\pm$0.43}\\
\hline
AIRL 	& -15.79$\pm$4.25 	& -28.75$\pm$1.07\\
\hline
WAIRL 	& -8.19$\pm$4.54 	& -13.81$\pm$1.19\\
\hline
AIRL state-only 	& 111.71$\pm$18.35 	& -8.87$\pm$0.79\\
\hline
WAIRL state-only 	& \textbf{298.09$\pm$42.18} 	& -8.82$\pm$0.20\\

\end{tabular}
\caption{\small Comparison of the GAIL, AIRL, GAIL and  WAIRL with  transfer learning tasks.}
\label{tab:transfer-tasks}
\end{table}

 In Figure \ref{fig:mu} we show heatmaps of the weights $\mu(s)$ obtained from the WAIRL-StateOnly algorithm with the Point Mass-Maze task, noting that visualizing the weights for Ant-Disabled task is not straightforward. The weights around the targets (i.e., the star on top of the square) seem to be higher than those in the other areas. As stated above, higher weights will yield more stochastic state-policies, implying that the policies in high-reward areas seem more stochastic than those in low-reward areas. 
 The heatmaps from the training and transfer tests are remarkably different which would possibly be due to the effect of the  transferred environment.
 The distributions of the weights shown in Figure \ref{fig:mu} and the superior of the weighted algorithms may explain why the stochasticity should be different over states to have better learning outcomes.

\textbf{Other imitation learning tasks.} We also evaluate our weighted algorithms on other high-dimensional imitation learning tasks. In this experiments, we do not test with transferred environments but  evaluate the algorithms with the same environments that they were trained. Table \ref{tab:mujoco} reports comparison results for three MuJoCo continuous control tasks \citep{Todorov2012mujoco}, showing that all algorithms perform similarly for Reacher-v2, but for the other tasks,  WAIRL and WGAIL outperform  their counterparts by a wide margin.

\begin{table}[htb]\small
\centering
\begin{tabular}{l|l|l|l}
Env.  & \begin{tabular}[c]{@{}l@{}}Reacher\\-v2\end{tabular} & \begin{tabular}[c]{@{}l@{}}Walker2d\\-v2\end{tabular} & \begin{tabular}[c]{@{}l@{}}Humanoid\\Standup-v2\end{tabular}  \\ 
\hline
GAIL  & -7.60$\pm$1.98\textasciitilde{}                      & 288.15$\pm$48.59\textasciitilde{}                     & 102403$\pm$16027\textasciitilde{}                             \\
WGAIL & -5.93$\pm$0.87\textasciitilde{}                      & 944.96$\pm$568.05\textasciitilde{}                    & 105345$\pm$10087\textasciitilde{}                             \\
AIRL  & -5.23$\pm$0.83\textasciitilde{}                      & 157.74$\pm$22.46\textasciitilde{}                     & 79120$\pm$27524\textasciitilde{}                              \\
WAIRL & \textbf{-5.18$\pm$0.65 }                             & \textbf{1562.1$\pm$945.3}                             & \textbf{143589$\pm$9298 }                                    
\end{tabular}
\caption{\small Comparison of the GAIL, AIRL, WGAIL and WAIRL on Mujoco tasks.}
\label{tab:mujoco}
\end{table}

\begin{figure}[htb] 
\centering
    \includegraphics[width=0.9\linewidth]{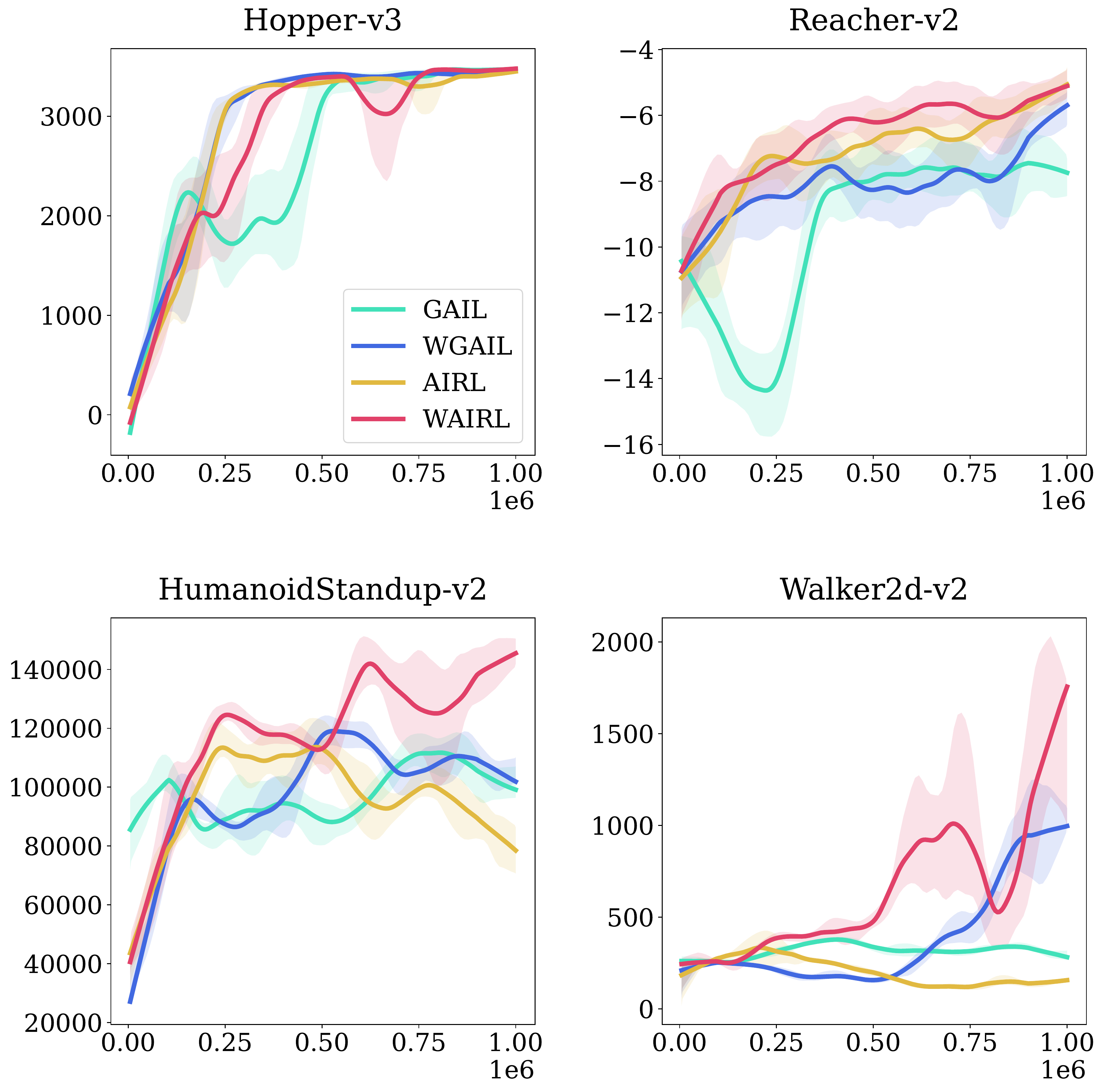} 
    \caption{\small  Comparison of the GAIL, WGAIL, AIRL and WAIRL on three MuJoCo environments, training for one million time-steps.  }
    \label{fig:performance-curves}
\end{figure}

 We also report the performance curves in Fig. \ref{fig:performance-curves} to report  how the algorithms run over time steps. The figure shows that our weighted algorithms are more stable  and have better performance than the other algorithms.

\section{Conclusion}
\label{sec:conclude}
We introduced the weighted  entropy framework for IRL and IM. Our framework allows us to learn, in addition to the expert's reward or policy function, the structure of the entropy function. We showed that, by learning the weights, we can recover the randomness/stochasticity of the  expert's policy. The learned reward and weight functions are portable, in the sense that they can be used to predict policies consistent with the expert on new  state space in the domain of the feature information. Our experiments on simulated and real-life environments, with both discrete and continuous control  tasks, showed that our approach outperforms prior methods that only return a reward or policy function.    

In this paper we assumed that there is  only one weight function to be learned, admitting that expert demonstrations may come from heterogeneous experts and there should be more than one weight function to be learned. This would be a direction for future work. Multi-agent IRL and IM with structural entropy or with a generalized regularizer would also be interesting directions to explore. 

\section{Acknowledgments}
This research/project is supported by the Singapore
Ministry of Education (MOE) Academic Research Fund
(AcRF) Tier 1 grant (Grant No: 20-C220-SMU-010).
\nocite{langley00}

\bibliography{refs}

\begin{thebibliography}{24}
\providecommand{\natexlab}[1]{#1}

\bibitem[{Abbeel and Ng(2004)}]{abbeel2004apprenticeship}
Abbeel, P.; and Ng, A.~Y. 2004.
\newblock Apprenticeship learning via inverse reinforcement learning.
\newblock In \emph{Proceedings of the twenty-first international conference on
  Machine learning}, 1. ACM.

\bibitem[{Arnob(2020)}]{Arnob2020off}
Arnob, S.~Y. 2020.
\newblock Off-Policy Adversarial Inverse Reinforcement Learning.
\newblock \emph{arXiv preprint arXiv:2005.01138}.

\bibitem[{Bagnell et~al.(2007)Bagnell, Chestnutt, Bradley, and
  Ratliff}]{Bagnell2007boosting}
Bagnell, J.; Chestnutt, J.; Bradley, D.~M.; and Ratliff, N.~D. 2007.
\newblock Boosting structured prediction for imitation learning.
\newblock In \emph{Advances in Neural Information Processing Systems},
  1153--1160.

\bibitem[{Bloem and Bambos(2014)}]{Bloem2014infinite}
Bloem, M.; and Bambos, N. 2014.
\newblock Infinite time horizon maximum causal entropy inverse reinforcement
  learning.
\newblock In \emph{53rd IEEE Conference on Decision and Control}, 4911--4916.
  IEEE.

\bibitem[{Finn et~al.(2016)Finn, Christiano, Abbeel, and
  Levine}]{Finn2016_connectionIRL}
Finn, C.; Christiano, P.; Abbeel, P.; and Levine, S. 2016.
\newblock A connection between generative adversarial networks, inverse
  reinforcement learning, and energy-based models.
\newblock \emph{arXiv preprint arXiv:1611.03852}.

\bibitem[{Finn, Levine, and Abbeel(2016)}]{Finn2016Guided}
Finn, C.; Levine, S.; and Abbeel, P. 2016.
\newblock Guided cost learning: Deep inverse optimal control via policy
  optimization.
\newblock In \emph{International Conference on Machine Learning}, 49--58.

\bibitem[{Fosgerau, Frejinger, and Karlstrom(2013)}]{Fosgerau2013_RL}
Fosgerau, M.; Frejinger, E.; and Karlstrom, A. 2013.
\newblock A link based network route choice model with unrestricted choice set.
\newblock \emph{Transportation Research Part B: Methodological}, 56: 70--80.

\bibitem[{Fu, Luo, and Levine(2017)}]{Fu2017Robust_IRL}
Fu, J.; Luo, K.; and Levine, S. 2017.
\newblock Learning robust rewards with adversarial inverse reinforcement
  learning.
\newblock \emph{arXiv preprint arXiv:1710.11248}.

\bibitem[{Geist, Scherrer, and Pietquin(2019)}]{Geist2019theory}
Geist, M.; Scherrer, B.; and Pietquin, O. 2019.
\newblock A theory of regularized markov decision processes.
\newblock \emph{arXiv preprint arXiv:1901.11275}.

\bibitem[{Goodfellow et~al.(2014)Goodfellow, Pouget-Abadie, Mirza, Xu,
  Warde-Farley, Ozair, Courville, and Bengio}]{Goodfellow2014GAN}
Goodfellow, I.; Pouget-Abadie, J.; Mirza, M.; Xu, B.; Warde-Farley, D.; Ozair,
  S.; Courville, A.; and Bengio, Y. 2014.
\newblock Generative adversarial nets.
\newblock In \emph{Advances in neural information processing systems},
  2672--2680.

\bibitem[{Ho and Ermon(2016)}]{Ho2016GAN_IRL}
Ho, J.; and Ermon, S. 2016.
\newblock Generative adversarial imitation learning.
\newblock In \emph{Advances in neural information processing systems},
  4565--4573.

\bibitem[{Jin et~al.(2015)Jin, Damianou, Abbeel, and Spanos}]{Jin2015inverse}
Jin, M.; Damianou, A.; Abbeel, P.; and Spanos, C. 2015.
\newblock Inverse reinforcement learning via deep gaussian process.
\newblock \emph{arXiv preprint arXiv:1512.08065}.

\bibitem[{Levine, Popovic, and Koltun(2010)}]{Levine2010feature}
Levine, S.; Popovic, Z.; and Koltun, V. 2010.
\newblock Feature construction for inverse reinforcement learning.
\newblock In \emph{Advances in Neural Information Processing Systems},
  1342--1350.

\bibitem[{Levine, Popovic, and Koltun(2011)}]{Levine2011nonlinearIRL}
Levine, S.; Popovic, Z.; and Koltun, V. 2011.
\newblock Nonlinear inverse reinforcement learning with Gaussian processes.
\newblock In \emph{Advances in Neural Information Processing Systems}, 19--27.

\bibitem[{Mai, Fosgerau, and Frejinger(2015)}]{Mai2015_nestedRL}
Mai, T.; Fosgerau, M.; and Frejinger, E. 2015.
\newblock A nested recursive logit model for route choice analysis.
\newblock \emph{Transportation Research Part B: Methodological}, 75: 100--112.

\bibitem[{Rasmussen(2003)}]{Rasmussen2003gaussian}
Rasmussen, C.~E. 2003.
\newblock Gaussian processes in machine learning.
\newblock In \emph{Summer School on Machine Learning}, 63--71. Springer.

\bibitem[{Ratliff, Bagnell, and Zinkevich(2006)}]{Ratliff2006maximum}
Ratliff, N.~D.; Bagnell, J.~A.; and Zinkevich, M.~A. 2006.
\newblock Maximum margin planning.
\newblock In \emph{Proceedings of the 23rd international conference on Machine
  learning}, 729--736.

\bibitem[{Ratliff, Silver, and Bagnell(2009)}]{ratliff2009learning}
Ratliff, N.~D.; Silver, D.; and Bagnell, J.~A. 2009.
\newblock Learning to search: Functional gradient techniques for imitation
  learning.
\newblock \emph{Autonomous Robots}, 27(1): 25--53.

\bibitem[{Russell(1998)}]{russell1998learning}
Russell, S.~J. 1998.
\newblock Learning agents for uncertain environments.
\newblock In \emph{COLT}, volume~98, 101--103.

\bibitem[{Syed and Schapire(2008)}]{syed2008game}
Syed, U.; and Schapire, R.~E. 2008.
\newblock A game-theoretic approach to apprenticeship learning.
\newblock In \emph{Advances in neural information processing systems},
  1449--1456.

\bibitem[{Todorov, Erez, and Tassa(2012)}]{Todorov2012mujoco}
Todorov, E.; Erez, T.; and Tassa, Y. 2012.
\newblock Mujoco: A physics engine for model-based control.
\newblock In \emph{2012 IEEE/RSJ International Conference on Intelligent Robots
  and Systems}, 5026--5033. IEEE.

\bibitem[{Yu, Song, and Ermon(2019)}]{Yu2019multi}
Yu, L.; Song, J.; and Ermon, S. 2019.
\newblock Multi-agent adversarial inverse reinforcement learning.
\newblock In \emph{International Conference on Machine Learning}, 7194--7201.
  PMLR.

\bibitem[{Ziebart, Bagnell, and Dey(2010)}]{ziebart2010_IRL_Causal}
Ziebart, B.~D.; Bagnell, J.~A.; and Dey, A.~K. 2010.
\newblock Modeling interaction via the principle of maximum causal entropy.
\newblock In \emph{Proceedings of the Twenty-seventh International Conference
  on Machine Learning (ICML'10)}, 1255--1262.

\bibitem[{Ziebart et~al.(2008)Ziebart, Maas, Bagnell, and
  Dey}]{Ziebart2008maximum}
Ziebart, B.~D.; Maas, A.~L.; Bagnell, J.~A.; and Dey, A.~K. 2008.
\newblock Maximum entropy inverse reinforcement learning.
\newblock In \emph{Proceedings of The Twenty-third AAAI Conference on
  Artificial Intelligence (AAAI'08)}, volume~8, 1433--1438. Chicago, IL, USA.

\end{thebibliography}

\newpage
\appendix
\onecolumn

\begin{center}
\textbf{\huge Supplementary Material}
\end{center}

\section{Proof of Proposition \ref{prop:prop-3}}
We provide a proof for Proposition \ref{prop:prop-3}, noting that Proposition \ref{prop:p1}  is just a special case where $\mu(s) = \eta$ for all $s\in \cS$. We write the Markov problem under our weighted causal entropy framework as
\begin{align}
&\max_{\substack{\bpii}}\Bigg\{\bbE_{\tau \sim \bpii}\Bigg[\sum_{t=0}^{\infty}\gamma^{t} \Big( r(a_t|s_t)  - \mu(s_t)\ln(\pi(a_t|s_t)) \Big)\Bigg]\Bigg\} \nonumber \\
\Leftrightarrow & \max_{\substack{\bpii}}\Bigg\{\bbE_{\tau \sim \bpii}\Bigg[\sum_{t=0}^{\infty}\gamma^{t} \Big( r(a_t|s_t)\Big)\Bigg]- \bbE_{\tau \sim \bpii}\Bigg[\sum_{t=0}^{\infty}\gamma^{t}  \mu(s_t)\left(\sum_{a\in\cA}\pi(a|s_t) \ln \pi(a|s_t)) \right)\Bigg]\Bigg\} \nonumber\\
\Leftrightarrow & \max_{\substack{\bpii}}\Bigg\{\bbE_{\tau \sim \bpii}\Bigg[\sum_{t=0}^{\infty}\gamma^{t} \Big( r(a_t|s_t)+ \left(\ln |\cA|\right) \mu(s_t)\Big)\Bigg]- \bbE_{\tau \sim \bpii}\Bigg[\sum_{t=0}^{\infty}\gamma^{t}  \mu(s_t)\KL\left(\pi(\cdot|s_t)\bigg|\bigg|\frac{\be}{|\cA|}\right)\Bigg]  \Bigg\}. \label{eq:proof-eq1}
\end{align}
For notational simplicity, let us denote 
$$
\begin{aligned}
f(\bpii) = \bbE_{\tau \sim \bpii}\Bigg[\sum_{t=0}^{\infty}\gamma^{t} \Big( r'(a_t|s_t)\Big)\Bigg];\; 
g(\bpii) =  \bbE_{\tau \sim \bpii}\Bigg[\sum_{t=0}^{\infty}\gamma^{t}  \mu(s_t)\KL\left(\pi(a|s_t)\bigg|\bigg|\frac{\be}{|\cA|}\right)\Bigg]    
\end{aligned}
$$
where $r'(a_t|s_t) = r(a_t|s_t) + \ln|\cA| \mu(s_t)$. We also let  $\bpii^*$ be an  optimal solution the Markov problem \eqref{eq:proof-eq1} and 
$
\alpha = g(\bpii^*).
$ We will prove that $\bpii^*$ is also optimal to the problem $\underset{\bpii}{\text{max}}\;  \{f(\bpii)|\; 
\text{s.t.} \;  g(\bpii) \leq \alpha\} \geq 0.$
By contradiction, let assume that there is an optimal  policy $\overline{\bpii}$ to this problem such that $f(\overline{\bpii}) >f(\bpii^*)$. We then have the following inequalities
\begin{align}
    f(\overline{\bpii}) &>f(\bpii^*) \nonumber \\
    g(\overline{\bpii}) &\leq \alpha = g(\bpii^*).\nonumber
\end{align}
Thus, $f(\overline{\bpii})+ g(\overline{\bpii}) > f(\bpii^*) + \alpha = f(\bpii^*) + g(\bpii^*)$, which is contrary to the initial assumption that $\bpii^*$ is optimal to the Markov problem $\max_{\bpii} f(\bpii) + g(\bpii)$. So, $\bpii^*$ is also optimal to the constrained problem $\underset{\bpii}{\text{max}}\;  \{f(\bpii)|\; 
\text{s.t.} \;  g(\bpii) \leq \alpha\}$ as desired. 

If $\mu(s_t) = \eta$ for all $s_t\in \cS$ as in the case of Proposition \ref{prop:prop-3}, then we can write 
\[
\begin{aligned}
f(\bpii) &=  \bbE_{\tau \sim \bpii}\Bigg[\sum_{t=0}^{\infty}\gamma^{t} \Big( r(a_t|s_t)\Big)\Bigg] + \ln|\cA|\eta \bbE_{\tau \sim \bpii} \left[\sum_{t=0}^\infty \gamma^t\right]\\
&=  \bbE_{\tau \sim \bpii}\Bigg[\sum_{t=0}^{\infty}\gamma^{t} \Big( r(a_t|s_t)\Big)\Bigg] + \frac{1}{1-\gamma}\ln|\cA|\eta.
\end{aligned}
\]
Then the term $\frac{1}{1-\gamma}\ln|\cA|\eta$ is a constant and can be removed from the constrained MDP optimization problem. We obtain the claim in Proposition \ref{prop:prop-3}. 

\section{Proof of Theorem \ref{th:optimal-policy}}
From the regularized MDP framework established in \cite{Geist2019theory}, we know that the value function $V^*$ is a solution to the contraction mapping $\cT[V] = V$, where $\cT[V]$ is defined as
\[
\cT[V] = \max_{\pi(\cdot|s)}\left\{ \bbE\left[r(a|s) + \gamma \sum_{s'}q(a'|a,s) V(s') \right] + \mu(s) \bbE[\ln \pi(a|s)]\right\} 
\]
The contraction property guarantees the existence and uniqueness of $V^*$ for any finite reward and weight functions $r(a|s)$ and  $\mu(s)$.  
The value of $\cT[V]$ for a given $V \in \bbR^{|\cS|}$ can be computed by solving
\begin{align}
\cT[V] = \underset{\pi(a|s),a\in\cA}{\text{max}}\qquad & \sum_{a\in\cA} \pi(a|s) Q(s,a,V) + \mu(s)\sum_{a\in\cA}  \pi(a|s)\ln \pi(a|s)& \nonumber \\
\text{s.t.} \qquad &   \sum_{a\in\cA} \pi(a|s) = 1& \nonumber\\
 &   \pi(a|s) \geq 0&\forall  a\in \cA, \nonumber
\end{align}
where $Q(s,a,V) = r(a|s) + \gamma \sum_{s'}q(a'|a,s) V(s')$.
So, if we divide the objective function by $\mu(s)$, then the above problem becomes a standard entropy-regularized one.
\begin{align}
\cT[V]/\mu(s) = \underset{\pi(a|s),a\in\cA}{\text{max}}\qquad & \sum_{a\in\cA} \pi(a|s) \frac{Q(s,a,V)}{\mu(s)} + \sum_{a\in\cA}  \pi(a|s)\ln \pi(a|s)& \nonumber \\
\text{s.t.} \qquad &   \sum_{a\in\cA} \pi(a|s) = 1& \nonumber\\
 &   \pi(a|s) \geq 0&\forall  a\in \cA, \nonumber
\end{align}
which yields an optimal solution as 
\begin{equation}
\label{eq:th3-eq1}
\pi^*(a|s) = \frac{\exp(Q(s,a,V)/\mu(s))}{\exp(\cT[V]/\mu(s))}    
\end{equation}
and the objective value
\begin{equation}
\label{eq:th3-eq2}
\cT[V] =\mu(s) \ln \Big(\sum_{a\in\cA}\exp(Q(s,a,V)/\mu(s))\Big).    
\end{equation}
Substitute $V^*$ into \eqref{eq:th3-eq1} and \eqref{eq:th3-eq2} we obtain desired results.

\section{First-order Derivatives of the Log-likelihood Function}
We discuss how to compute the log-likelihood function and its gradients for our weighted MLE-based IRL algorithms. For the generative adversarial algorithms, the gradients of the discriminators and policy functions are straightforward.

We write the log-likelihood function given in (7) as
\[
\cL(\cD^E|\bpsi,\btheta) = \sum_{\tau\in \cD^E} \sum_{(a,s)\in\tau } \frac{r_\btheta (a|s) +\gamma \sum_{s'} q(s'|s,a) V^{\bpsi,\btheta}(s') - V^{\bpsi,\btheta}(s)}{\mu_{\bpsi}(s)}.
\]
Taking the first-order derivatives of $\cL(\cD^E|\bpsi,\btheta)$ w.r.t a parameter $\psi_i$ or $\theta_i$ gives
\begin{align}
    \frac{\partial \cL(\cD^E|\bpsi,\btheta)}{\partial \theta_i} &= \sum_{\tau\in \cD^E} \sum_{(a,s)\in\tau } \frac{\Delta^r_{\theta_i} (a|s) +\gamma \sum_{s'} q(s'|s,a) \Delta^V_{\theta_i}(s') - \Delta^V_{\theta_i}(s)}{\mu_{\bpsi}(s)}\nonumber \\
        \frac{\partial \cL(\cD^E|\bpsi,\btheta)}{\partial \psi_i} &= \sum_{\tau\in \cD^E} \sum_{(a,s)\in\tau } \frac{ \gamma \sum_{s'} q(s'|s,a) \Delta^V_{\psi_i}(s') - \Delta^V_{\psi_i}(s)}{\mu_{\bpsi}(s)}   \nonumber\\
        &\qquad - \frac{(r_\btheta (a|s)
         +\gamma \sum_{s'} q(s'|s,a) V^{\bpsi,\btheta}(s') - V^{\bpsi,\btheta}(s)) \Delta^\mu_{\psi_i}(s)}{(\mu_\bpsi (s))^2}, \nonumber 
\end{align}
where $\Delta^V_{\theta_i}$, $\Delta^V_{\psi_i}$ are the partial derivatives of $V^{\bpsi,\btheta}$ w.r.t parameters $\theta_i$ and $\psi_i$, respectively, and $\Delta^{r}_{\theta_i}$, $\Delta^{\mu}_{\psi_i}$ are the partial derivatives of $r_{\btheta}(a|s)$ and $\mu_{\bpsi}(s)$ w.r.t parameters $\theta_i$ and $\psi_i$, respectively. So, in order to compute the derivatives of the log-likelihood function, we need to compute $\Delta^V_{\theta_i}$, $\Delta^V_{\psi_i}$. This can be done by taking the derivatives of the mapping $\cT[V]$ w.r.t $\theta_i$ and $\psi_i$ as
\begin{align}
    \frac{\partial \cT[V](s)}{\partial \theta_i} &= \mu_{\bpsi}(s) \frac{\sum_{a\in\cA} \frac{1}{\mu_{\bpsi}(s)}\exp(Q(a,s,V)/\mu_{\bpsi}(s)) \left(  \Delta^r_{\theta_i} + \gamma \bbE_{s'}\left[\frac{\partial V(s')}{\partial \theta_i}\right]\right)
    }{\exp(\cT[V](s)/\mu_{\bpsi}(s))} \label{eq:th1:eq3} \\
    \frac{\partial \cT[V](s)}{\partial \psi_i} &= \Delta^{\mu}_{\psi_i}(s) \frac{\cT[V](s)}{\mu_{\bpsi}(s)} + \mu_{\bpsi}(s) \frac{\sum_{a\in\cA} \exp(Q(a,s,V)/\mu_{\bpsi}(s))U^a_{\psi_i}
    }{\exp(\cT[V](s)/\mu_{\bpsi}(s))} \label{eq:th1:eq4}
\end{align}
where
\[
U^a_{\psi_i} = \frac{1}{\mu_{\bpsi}(s)}\left( \gamma \bbE_{s'|s,a}\left[\frac{\partial V(s')}{\partial \psi_i}\right]\right) - Q(a,s,V)\frac{\Delta^{\mu}_{\psi_i}(s)}{(\mu_{\bpsi}(s))^2}.
\]
The derivatives of the value function can be computed simultaneously with the computation of the value function. The following algorithm describes detailed steps to do this.
\begin{algorithm}[htb]
\caption{\textit{Log-likelihood and gradient computation}}
\label{algo:LL-grad}
\begin{algorithmic}[1]
\STATE Set $V = 0$; $\partial V/\partial \theta_i = 0;$ $\partial V/\partial \psi_i = 0$; and an accuracy threshold $\epsilon>0$\\
\REPEAT 
\STATE Compute:
\begin{align}
 V &\leftarrow \cT[V] ;\nonumber \\
 \frac{\partial V}{\partial \theta_i} &\leftarrow \frac{\partial \cT[V]}{\partial \theta_i}  ;\;
  \frac{\partial V}{\partial \psi_i} \leftarrow \frac{\partial \cT[V]}{\partial \psi_i}  \nonumber 
\end{align}
\UNTIL {$||V-\cT[V]|| \leq \epsilon$}
\STATE Return $V^* = V; \Delta^V_{\theta_i} = \partial V/\partial \theta_i;
\; \Delta^V_{\psi_i} = \partial V/\partial \psi_i$
\end{algorithmic}
\end{algorithm}

\section{{Experiments with Highway Driving Simulator}}
\begin{figure}[htb] 
\centering
    \includegraphics[width=0.9\linewidth]{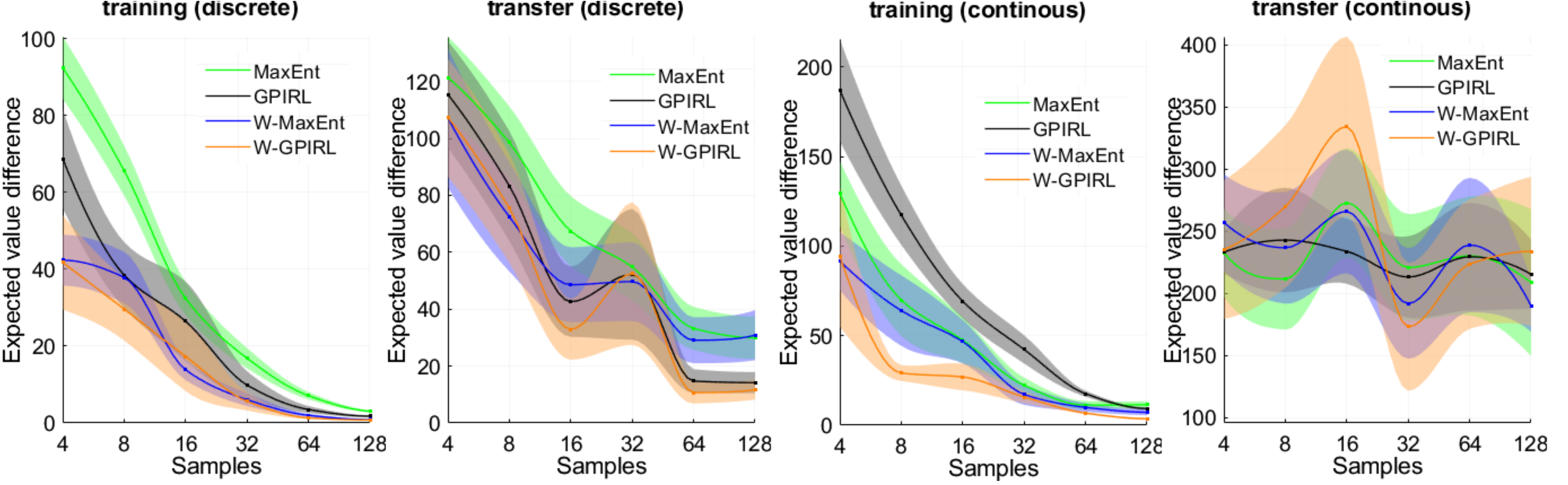} 
    \caption{\small Highway driving behavior experiments, solid curves show the means and shading show standard errors.} 
    \label{fig:hig_expected_value} 
\end{figure}

We also evaluate our algorithm with a simple highway driving simulator \citep{Levine2010feature,Levine2011nonlinearIRL}. The task is to navigate a vehicle in a highway of three lanes with all vehicles moving with the same speed.
The MDP is deterministic in the context.
The true rewards are constructed in such a way that the agent should drive as fast as possible, but avoid driving more than double the speed of the traffic within two-car length of a police vehicle. There are features indicating the distance  to the nearest vehicle of  a specific class (car or motorcycle) or category (civilian or police)
in front of the agent, either in the same of a different lane. 
In analogy to  the objectworld experiments, we also discretize the continuous features. 
In this context, the true rewards become highly nonlinear with respect to the feature information, making them challenging to be recovered for both the MaxEnt and GP-based algorithms. We generate demonstrations  of length 32  and test the IRL algorithms with varying numbers of samples, from 4 to 128. We plot the means  and standard errors of the expected value differences in Fig. \ref{fig:hig_expected_value}. The MaxEnt and GPIRL were consistently outperformed by our algorithms on the training environments, especially with small sample sizes. On the transfer environments, our weighted versions are still better in the case of discrete features, but the improvement is only modest. For the case of continuous features, all the four methods are comparable. It is interesting to see that the W-MaxEnt generally outperforms the GPIRL, indicating the ability of our weighted framework to enhance linear methods (i.e., MaxEnt) recovering optimal policies from highly nonlinear true rewards.

\end{document}